\documentclass[runningheads]{llncs}

\usepackage{marvosym} 
\usepackage{cite}
\usepackage{amsmath,amssymb,amsfonts}
\usepackage{fancyhdr}
\usepackage{graphicx}
\usepackage{textcomp}
\usepackage{xcolor}
\usepackage{booktabs}
\usepackage{multirow}
\usepackage{algorithm}
\usepackage{algpseudocode}
\usepackage{url}
\usepackage{tikz}
\usetikzlibrary{shapes.geometric, arrows.meta, fit, backgrounds}
\usepackage[most]{tcolorbox}

\newtcolorbox{keyinsightbox}[1][]{
  colback=blue!4,
  colframe=blue!60!black,
  fonttitle=\bfseries,
  arc=2pt,
  boxrule=0.8pt,
  left=6pt, right=6pt, top=4pt, bottom=4pt,
  title=#1
}

\def\BibTeX{{\rm B\kern-.05em{\sc i\kern-.025em b}\kern-.08em
    T\kern-.1667em\lower.7ex\hbox{E}\kern-.125emX}}

\begin{document}

\title{MOSCOPT: Mixture-of-Skills Collective Optimization for LLM Agents }

\author{
  Zhenyu Zhang\inst{1}\textsuperscript{\Letter} \and
  Jiudong Yang\inst{2}
}

\authorrunning{Z. Zhang and J. Yang}

\institute{
  ML Center, Coupang, Shanghai, China
  \\
  \email{zhangzhenyu13@outlook.com} \and
  AI Center, Futu AI, Shenzhen, China
}

\maketitle

\begin{abstract}
Natural language prompts and skills serve as the strategic backbone of LLM-based agents. Recent advances in prompt and skill optimization have achieved notable gains, yet all existing methods optimize a \emph{single} text template---missing the synergy among multiple complementary strategies.
We propose MOSCOPT, a text-native, parameter-free algorithm that jointly optimizes a pool of $N$ skills and a gating skill $G$ that dynamically selects $K$ skills per step.
To effectively optimize the skills, we build the EditAdam with internally maintained dual states.
Through the three-phase interleaved updates with EditAdam, the system monotonically improves without gradient or parameter tuning.
Extensive experiments and detailed ablations across 5 benchmarks and 3 target LLMs demonstrate that MOSCOPT consistently outperforms all baselines, and confirm that both the mixture-of-skills architecture with selective activation and the collective evolution with three-phase interleaving are essential to its superior performance. Code is released \footnote{\url{https://github.com/zhangzhenyu13/SummerClaw/tree/master/summerclaw/agent_trainer/algorithms/moscopt}}.

\keywords{LLM agent, skill optimization, collective evolution}

\end{abstract}

\section{Introduction}
\label{sec:introduction}

Textual skills---natural language strategy documents injected into an agent's context---have emerged as a critical paradigm for deploying LLM-based agents~\cite{jiang2026sok}.
They encode reusable procedural knowledge such as planning heuristics, tool-use policies, and domain-specific rules, enabling agents to perform complex tasks without modifying model weights~\cite{sumers2024cognitive}.
However, real-world tasks often demand \emph{multiple complementary strategies}: a calculation sub-task may require a math skill, a search sub-task a retrieval skill, and a verification sub-task a checking skill---all within a single trajectory.
A single monolithic skill cannot simultaneously excel at these diverse roles, and the agent has no mechanism to dynamically invoke the right strategy at the right time.

Existing approaches to agent skill/prompts optimization fall into three categories, each with fundamental limitations.
\emph{Manual or one-shot skills} are labor-intensive and cannot adapt from execution feedback.
\emph{Self-improvement methods}~\cite{shinn2023reflexion} refine outputs through verbal self-reflection but operate on individual instances without producing persistent reusable skills.
\emph{Single-skill optimization}, exemplified by SkillOpt~\cite{yang2026skillopt}, treats a skill as a trainable text-space object with disciplined bounded editing and strict-improvement validation---yet optimizes only a \emph{single} skill, making it impossible to model strategy interactions or dynamic skill combination.
Beyond these three categories, population-level methods~\cite{lau2025dipper} and evolutionary coordination~\cite{dang2025evolving} evolve multiple prompts or agents, but target independent instances rather than co-existing strategies within a single agent's context. 
Meanwhile, mixture-of-experts (MoE) routing~\cite{ yue2025masrouter} demonstrates the power of dynamic expert selection, but relies on parameterized neural routers that are neither interpretable nor editable as text.

To bridge these gaps, we propose \textbf{MOSCOPT}, the first fully text-native, parameter-free framework for \emph{joint optimization of a skill pool and a gating scheduler}.
MOSCOPT addresses the three core challenges as follows:
\textbf{Mixture-of-Skills.} Rather than a single skill, MOSCOPT maintains a pool of $N$ complementary skills and a gating skill $G$ that dynamically activates exactly $K$ skills per task or execution step via progressive disclosure of skill summaries, while non-activated skills remain invisible to prevent context overload.
\textbf{EditAdam based Collective Evolution.} We build EditAdam to optimize the skill pool with three-phase interleaved update scheme including skill editing, gating editing, and collective evolution.
\textbf{Text-Native Routing with Multi-Mode Deployment.} Unlike neural MoE routers, the gating skill is a text document optimized through the same bounded-edit pipeline as the skills it schedules. The system supports flexible deployment via full mixture gating, distilled single skill, or static routing table extraction. Our evaluation further confirms the effectiveness of the proposed method.


\section{Related Work}
\label{sec:relatedworks}


\textbf{Agent Skill Optimization}:
Recent surveys~\cite{jiang2026sok} characterize agent skills as reusable procedural knowledge that goes beyond tool use, encompassing planning heuristics, domain rules, and interaction protocols.
Complementary work on agent memory---including procedural memory extraction~\cite{lin2025seagent} and agentic memory systems~\cite{xu2025amem}---shows that persistent, structured knowledge improves agent performance across tasks.
OPRO~\cite{yang2024opro} uses LLMs themselves to search over prompt space, while self-improvement approaches such as Reflexion~\cite{shinn2023reflexion} enable agents to learn from failures through verbal self-reflection.
SkillOpt~\cite{yang2026skillopt} advances this line by treating a skill as a trainable external state with disciplined text-space optimization,.
However, all these methods optimize a \emph{single} skill or prompt, which we identify as a fundamental limitation for tasks requiring diverse, complementary strategies.
\textbf{Collective Evolution}:
EvoSkill~\cite{alzubi2026evoskill} discovers skills through evolutionary optimization.
At the population level, PromptBreeder~\cite{fernando2023promptbreeder} evolves a population of task-prompts via self-referential mutation.
Multi-agent systems have adopted evolutionary coordination, evolving orchestration~\cite{dang2025evolving} dynamically adapts collaboration strategies across agents. 
However, these methods evolve individual components without modeling the \emph{interaction} among multiple co-existing strategies within a single agent's skill pool.
\textbf{Routing and Mixture Paradigms}:
The mixture-of-experts (MoE) paradigm~\cite{he2024mixture} addresses specialization through learned routing among expert networks.
However, MoE routing remains parameterized within neural architectures, lacking the interpretability and editability of text-based strategies.

\paragraph{Positioning.}
MOSCOPT differs from the above works in three key dimensions:
(1)~it maintains and optimizes a \emph{population} of skills, unlike single-skill optimization;
(2)~collective optimization operates on a \emph{single agent's skill pool} with synergistic credit assignment, unlike multi-agent evolutionary coordination;
(3)~the gating skill $G$ is a \emph{textual router} optimized through the same bounded-edit pipeline as the skills it schedules---parameter-free, interpretable, and protected by progressive disclosure---unlike neural MoE routing.

\section{Method: MOSCOPT}
\label{sec:method}

The core insight of MOSCOPT is that the gating mechanism is itself treated as a natural language skill---a text document optimized through trajectory feedback rather than gradient descent---unifying routing and skill optimization into a single parameter-free framework.

\subsection{Mixture-of-Skills Architecture}
\label{sec:mixture}
\begin{figure*}[t]
\centering
\includegraphics[width=\textwidth]{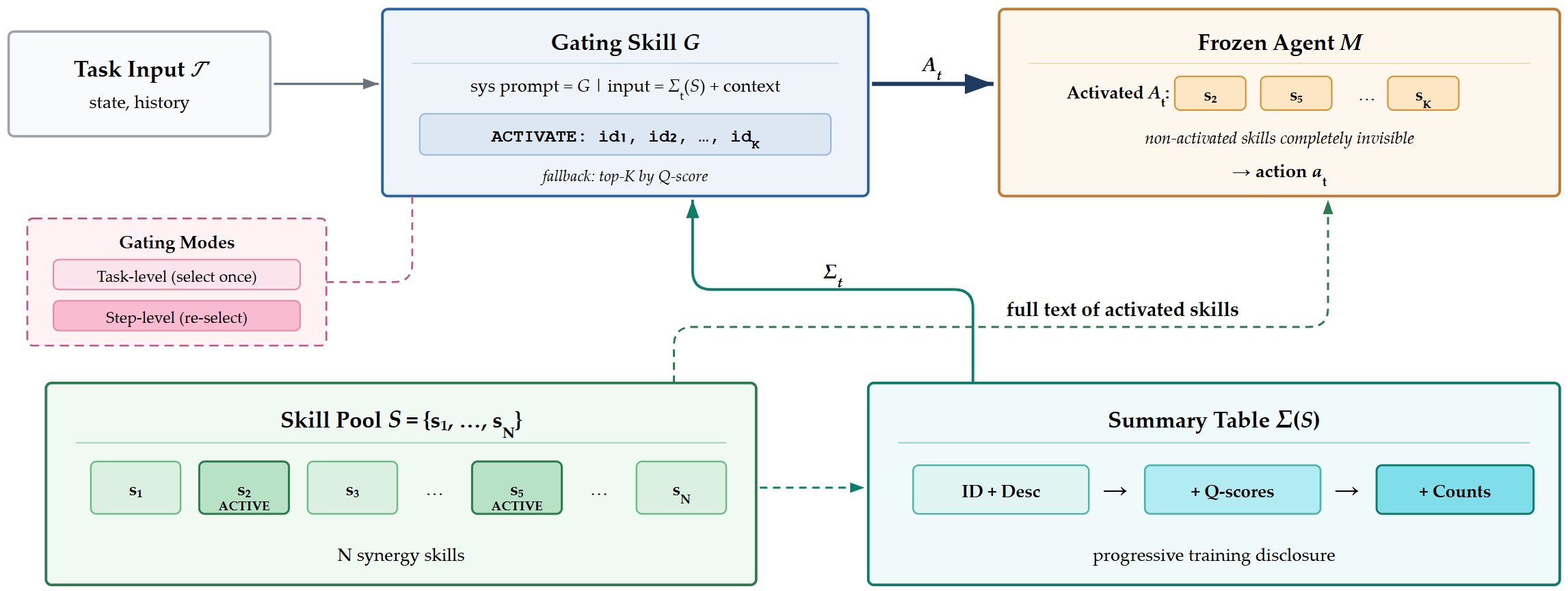}
\caption{
MOSCOPT maintains three core components: a \textbf{skill pool} $\mathcal{S} = \{s_1, \ldots, s_N\}$, a \textbf{gating skill} $G$, and a \textbf{summary table} $\Sigma(\mathcal{S})$.
Given a task input, $G$ selects $K$ skills via $\Sigma_t(\mathcal{S})$, outputting \texttt{ACTIVATE} IDs (with top-$K$-by-Q-score fallback).
The $K$ activated skills are loaded into agent context while non-activated skills remain invisible.
}
\label{fig:overview}
\end{figure*}

As illustrated in Figure~\ref{fig:overview}, given a task distribution $\mathcal{T}$ and a frozen base LLM agent $M$, MOSCOPT maintains three core components:
\textbf{skill pool} $\mathcal{S} = \{s_1, \ldots, s_N\}$, where each $s_i$ is a natural language strategy document.
\textbf{gating text skill} $G$, a natural language scheduler that selects exactly $K$ skills per task or execution step.
\textbf{summary table} $\Sigma(\mathcal{S})$ providing lightweight skill descriptions to $G$, following a progressive schedule: early epochs expose only skill IDs and descriptions; later epochs additionally reveal Q-scores and activation counts for score-informed selection.
The objective is to jointly optimize $\mathcal{S}$ and $G$ to maximize task success rate $\mathbb{E}_{\mathcal{T}}[R(\tau)]$, where $R(\tau)$ is task reward function.

\textbf{Execution with Gating.}
The gating skill $G$ serves as the system prompt of a dedicated LLM call. The user prompt concatenates the summary table $\Sigma_t(\mathcal{S})$, the current task state, and recent execution history. The LLM outputs exactly $K$ skill IDs in the format \texttt{ACTIVATE: id1, id2, \ldots}, parsed by a regular expression. On parse failure, the system falls back to top-$K$ selection by Q-score and records the failure as a future editing signal for $G$.
Once the activation set $\mathcal{A}_t$ ($|\mathcal{A}_t| = K$) is determined, the full text of the $K$ activated skills is loaded into the agent's context. Gating can operate at task level (select once per task) or step level (re-select at every execution step for long-horizon tasks).

\subsection{Collective Skill Evolution}

\begin{algorithm}[t]
\caption{MOSCOPT: Mixture-of-Skills Collective Optimization}
\label{alg:moscopt}
\begin{algorithmic}[1]
\Require Task distribution $\mathcal{T}$, base LLM $M$, pool size $N$, activation count $K$, evolution interval $E$
\Ensure Optimized skill pool $\mathcal{S}$, gating skill $G$
\State \textbf{Initialize:} $\mathcal{S} \leftarrow \{s_1,\ldots,s_N\}$, $G \leftarrow$ \textsc{GenerateGating}(), $\Sigma \leftarrow$ \textsc{Summarize}$(\mathcal{S})$, $Q[s_i] \leftarrow 0$ $\forall s_i$
\State $\mathit{opt} \leftarrow$ \textsc{EditAdam}($M$)
\For{\textbf{epoch} $= 1$ \textbf{to} $E_{\max}$}
  \State \Comment{\textit{Rollout with gating}}
  \State $\mathcal{D} \leftarrow$ \textsc{Rollout}($\mathcal{S}, G, \Sigma, K, \mathcal{T}_{\text{train}}, M$)
  \State Update $Q[s_i]$ via Eq.~\ref{eq:qscore}; update co-occurrence $C(i,j)$
  \State \Comment{\textit{Phase 1: Skill editing via EditAdam} ($G$ fixed)}
  \For{each $s_i \in$ \textsc{RankByFailure}($\mathcal{S}, \mathcal{D}$)}
    \State $\mathcal{S}[s_i] \leftarrow \mathit{opt}$.\textsc{step}($s_i, \mathcal{D}, \mathcal{T}_{\text{val}}, G$) \Comment{Alg.~\ref{alg:editadam}}
  \EndFor
  \State \Comment{\textit{Phase 2: Gating editing via EditAdam} ($\mathcal{S}$ fixed)}
  \State $G \leftarrow \mathit{opt}$.\textsc{step}($G, \mathcal{D}, \mathcal{T}_{\text{val}}, \mathcal{S}$) \Comment{Alg.~\ref{alg:editadam}}
  \State \Comment{\textit{Phase 3: Collective evolution} (every $E$ epochs)}
  \If{epoch $\bmod E = 0$}
    \State \textsc{Eliminate}($\mathcal{S}$, $Q$); \textsc{Reproduce}($\mathcal{S}$, $Q$)
    \State \textsc{CoBreed}($\mathcal{S}$, $C$); $\Sigma \leftarrow$ \textsc{RefreshSummary}($\mathcal{S}$)
  \EndIf
  \State $\mathit{opt}$.\textsc{memUpdate}() \Comment{2nd-order reflection inside EditAdam}
\EndFor
\State \textbf{return} $\mathcal{S}, G$
\end{algorithmic}
\end{algorithm}

\textbf{Credit Assignment.}
After each rollout batch, trajectory rewards are attributed to individual skills and the gating. For $K > 1$ co-activated skills, the binary reward $R(\tau) \in \{0,1\}$ is equally shared. For each skill $s_i$, the Q-score is updated via exponential moving average:
\begin{equation}
\label{eq:qscore}
Q_{\text{new}}(s_i) = (1 - \beta) \cdot Q_{\text{old}}(s_i) + \beta \cdot \frac{\sum_{\tau} c_i(\tau) \cdot R(\tau)}{\sum_{\tau} c_i(\tau)}
\end{equation}
where $c_i(\tau) \in \{0,1\}$ indicates whether $s_i$ was activated during rollout $\tau$, and $\beta = 0.3$. Additionally, when two skills $s_i, s_j$ are co-activated in a \emph{successful} trajectory ($R(\tau) = 1$), their co-occurrence count $C(i,j)$ is incremented. High-synergy pairs ($C(i,j)$ above threshold) are protected from elimination during collective evolution. 

\textbf{Three-Phase Interleaved Update.}
To prevent oscillation from simultaneous changes, MOSCOPT employs a three-phase update scheme over $E_{\max}$ epochs (Algorithm~\ref{alg:moscopt}). Each epoch runs on a training split $\mathcal{T}_{\text{train}}$; candidates are accepted only if they improve performance on a held-out validation split $\mathcal{T}_{\text{val}}$. The full optimization loop is given in Algorithm~\ref{alg:moscopt}; each skill and gating edit is delegated to the EditAdam optimizer (Algorithm~\ref{alg:editadam}):
\textbf{1 (Skill Editing):}
Each epoch begins with a \textsc{Rollout} that collects execution traces $\mathcal{D}$ over the training split $\mathcal{T}_{\text{train}}$. Skills are ranked by failure frequency via \textsc{RankByFailure}($\mathcal{S}, \mathcal{D}$). For each skill $s_i$, the optimizer invokes $\mathit{opt}$.\textsc{step}($s_i, \mathcal{D}, \mathcal{T}_{\text{val}}, G$), which internally performs bounded-edit proposal generation. 
\textbf{2 (Gating Editing):}
After skills are updated, the gating skill $G$ is edited through the same interface: $\mathit{opt}$.\textsc{step}($G, \mathcal{D}, \mathcal{T}_{\text{val}}, \mathcal{S}$). The optimizer uses selection error patterns (missed activations, poor combinations) from $\mathcal{D}$ to propose and validate edits, maintaining a separate step buffer for $G$.
\textbf{3 (Collective Evolution):}
Triggered every $E$ epochs at the pool level: \emph{eliminate} removes the lowest-Q-score skills (protecting insufficiently sampled and high-synergy pairs); \emph{reproduce} mutates top performers into new skills using their stable rules as genetic material; \emph{co-breed} merges pairs $(s_i, s_j)$ with exceptionally high synergy $C(i,j)$ into a unified skill; \textsc{RefreshSummary} then updates $\Sigma(\mathcal{S})$ to reflect the new pool composition.

\textbf{EditAdam memory update.}
At the end of each epoch, $\mathit{opt}$.\textsc{memUpdate}() is called. Inside EditAdam, the current epoch buffer $\mathcal{B}_{\text{curr}}$ (accumulated during \textsc{step} calls) is reflected against $\mathcal{B}_{\text{prev}}$ to update the long-term memory $\mathcal{M}$, then $\mathcal{B}_{\text{prev}} \leftarrow \mathcal{B}_{\text{curr}}$ and $\mathcal{B}_{\text{curr}}$ is cleared for the next epoch.

\subsection{EditAdam: Dual-State Text Optimizer}
\label{sec:editadam}

A distinguishing feature of MOSCOPT is that every update decision---to skills and to the gating skill---is produced by a text-native optimizer we call \textbf{EditAdam} (Algorithm~\ref{alg:editadam}). Unlike gradient-based optimizers that maintain numerical moments over parameters, EditAdam maintains \emph{two orders of textual state} over candidate edits, enabling stable, monotonic improvement without any learning rate or parameter tuning. The concurrent work SkillOpt leverages step + epoch feedbacks for bounded edits of single skill with distinguished different operations. However, EditAdam treats skill pool and gating skill as "unified optimizable parameters" in a unified framework.

\begin{algorithm}[t]
\caption{EditAdam: Dual-State Text Optimizer}
\label{alg:editadam}
\begin{algorithmic}[1]
\Require Base LLM $M$; per call: text unit $u$ (skill $s_i$ or gating $G$), execution traces $\mathcal{D}$, validation split $\mathcal{T}_{\text{val}}$, frozen context $C$ (either $G$ or $\mathcal{S}$)
\Ensure Updated text unit $u'$
\Statex \textit{\textbf{Internal state:}} base LLM $M$; \textbf{1st-order}: step buffer $\mathcal{B}_{\text{rej}}[u]$; \textbf{2nd-order}: long-term memory $\mathcal{M}$, epoch buffers $\mathcal{B}_{\text{curr}} \leftarrow \emptyset$, $\mathcal{B}_{\text{prev}} \leftarrow \emptyset$
\Statex
\Statex \textbf{procedure} \textsc{step}($u, \mathcal{D}, \mathcal{T}_{\text{val}}, C$):
\State $\{e_1, \ldots, e_m\} \leftarrow$ \textsc{EditProposal}($u, \mathcal{D}, \mathcal{B}_{\text{rej}}[u], \mathcal{M}$) \Comment{dual-state: 1st-order $\mathcal{B}_{\text{rej}}$ + 2nd-order $\mathcal{M}$ guide candidate edit proposal}
\For{$k = 1$ \textbf{to} $m$}
  \State $u_k' \leftarrow$ \textsc{BoundedEdit}($u, e_k, \mathcal{B}_{\text{rej}}[u], \mathcal{M}$)
  \If{\textsc{Score}($u_k'$, $C$, $\mathcal{T}_{\text{val}}$) $>$ \textsc{Score}($u$, $C$, $\mathcal{T}_{\text{val}}$)}
    \State $\mathcal{B}_{\text{rej}}[u] \leftarrow \emptyset$ \Comment{accept: clear step buffer}
    \State $\mathcal{B}_{\text{curr}}$.add($u$, \texttt{accept}, $e_k$, $\Delta$score) \Comment{accumulate into current epoch buffer}
    \State \Return $u_k'$
  \Else
    \State $\mathcal{B}_{\text{rej}}[u]$.add($e_k$, $\Delta$score) \Comment{record failure as 1st-order state}
    \State $\mathcal{B}_{\text{curr}}$.add($u$, \texttt{reject}, $e_k$, $\Delta$score)
  \EndIf
\EndFor
\State \Return $u$ \Comment{no edit accepted}
\Statex
\Statex \textbf{procedure} \textsc{memUpdate}():
\State $\mathcal{M} \leftarrow$ \textsc{Reflect}($\mathcal{M}, \mathcal{B}_{\text{curr}}, \mathcal{B}_{\text{prev}}$) \Comment{2nd-order: reflect on epoch-buffer change to update $\mathcal{M}$}
\State $\mathcal{B}_{\text{prev}} \leftarrow \mathcal{B}_{\text{curr}}$;\quad $\mathcal{B}_{\text{curr}} \leftarrow \emptyset$
\end{algorithmic}
\end{algorithm}

\textbf{Edit operations.}
Each edit proposal $e_k$ produced by \textsc{EditProposal} specifies one of two atomic text operations applied at a given position in $u$: \emph{add} (insert a text span at a specified location) or \emph{delete} (remove a contiguous span). \textsc{BoundedEdit} executes the proposed operation under the constraints of $\mathcal{B}_{\text{rej}}$ and $\mathcal{M}$, ensuring edits remain local and do not re-propose previously rejected directions.

\textbf{Dual-state hierarchy.}
EditAdam maintains two orders of textual state that jointly guide every \textsc{step} call. The \emph{first-order state} is the step buffer $\mathcal{B}_{\text{rej}}$, distilled by reflecting on execution trajectories: it records which edits were rejected and why, injected into the next \textsc{step} prompt to prevent re-proposing failed directions (text-space momentum). The \emph{second-order state} is the long-term memory $\mathcal{M}$, obtained by reflecting on the \emph{change} of epoch buffers across epochs---since each epoch buffer $\mathcal{B}_{\text{epoch}}$ is itself a summary of the first-order state, tracking how it evolves reveals durable optimization patterns (e.g., which edit strategies consistently improve scores). Both $\mathcal{B}_{\text{rej}}$ and $\mathcal{M}$ flow into \textsc{BoundedEdit}: the step buffer provides local, within-epoch guidance; $\mathcal{M}$ provides cross-epoch strategic priors, closing the loop from epoch-level reflection to step-level action.

\subsection{Implementation Designs}

\textbf{Deployment modes.}
The system supports three deployment modes: (i)~\emph{full mixture} ($\{\mathcal{S}, G, K\}$) with dynamic gating; (ii)~\emph{static routing table}, extracting gating Q score for fixed routing; (iii)~\emph{distilled single skill}, merging top skills into one prompt. We compare these modes in \S\ref{sec:discussion}.
\textbf{Preventing skill pool collapse.}
Repeated elimination/reproduction cycles and shared editing signals across tasks both risk driving skills toward homogeneous texts. Phase~3 mitigates this via: \emph{synergy protection} (high-$C(i,j)$ pairs survive elimination), \emph{mutation-based reproduction} (offspring explore distinct strategy regions), and \emph{forced diversity injection} when pairwise similarity exceeds $0.85$. This ensures the gating can make non-trivial selection decisions throughout training.

\section{Experimental Results}
\label{sec:evaluation}


\subsection{Setup}

\textbf{Benchmarks and Evaluations.}
We evaluate on five benchmarks: SearchQA~\cite{dunn2017searchqa}, GPQA-Diamond~\cite{rein2024gpqa}, HLE~\cite{phan2025hle}, AIME~\cite{aime2025}, and LiveMathematician~\cite{he2026livemath}.
For AIME, we pool problems from the 2024-2026 and split them into 50/10/30 for train/val/test with seed=42. We sample 100/20/50 for the rest. 
To ensure stable and efficient estimation, we adopt adaptive stopping rule with a minimum of 3 and maximum of 50 independent runs, stopping early when the sample standard deviation across completed runs falls below setting thresholds and report means. LLMs are invoked with temperature 0.8 and top-p 0.6.
\textbf{Agent Harness.}
We equips every agent with 3 tools: (i)~\emph{Shell}; (ii)~\emph{Python}; and (iii)~\emph{Browser Search}.
\textbf{Baselines.}
We compare 9 baselines:
(i)~\emph{Basic}: No Skill and 5-shot LLM Skill;
(ii)~\emph{Agent algorithm}: Reflexion~\cite{shinn2023reflexion}, DSPy~\cite{khattab2023dspy}, and SE-Agent~\cite{lin2025seagent};
(iii)~\emph{Skill algorithm}: OPRO~\cite{yang2024opro}, Trace2Skill~\cite{ni2026trace2skill}, EvoSkill~\cite{alzubi2026evoskill}, and SkillOpt~\cite{yang2026skillopt}.

\subsection{Overall Performance}

\begin{table*}[t]
\begin{center}
\small
\begin{tabular}{lcccccc}
\toprule
\textbf{Method} & \textbf{SearchQA} & \textbf{GPQA-D} & \textbf{HLE} & \textbf{AIME} & \textbf{LiveMath} & \textbf{avg.} \\
\midrule
No Skill & 48.8 & 90.7 & 46.3 & 94.5 & 29.9 & 62.0\\
5-shot LLM Skill & 41.0 & 88.3 & 46.5 & 93.1 & 29.4 & 59.7 \\
\midrule
Reflexion & 76.1 & 90.4 & \textbf{48.7} & 98.3 & 37.1 & 70.1 \\
DSPy & 62.2 & 90.7 & 46.1 & 93.8 & 31.5 & 64.9 \\
SE-Agent & 79.9 & 89.4 & 47.0 & 94.1 & 46.3 & 71.3 \\
\midrule
OPRO & 53.2 & 90.9 & 46.5 & 95.7 & 41.6 & 65.6 \\
Trace2Skill & 67.5 & 90.5 & 48.2 & 97.1 & 32.9 & 67.2 \\
EvoSkill & 77.4 & 91.5 & 48.1 & 96.5 & 46.8 & 72.1 \\
SkillOpt & 76.8 & 90.1 & 46.6 & 98.7 & 45.2 & 71.5\\
\midrule
\textbf{MOSCOPT} & \textbf{81.3} & \textbf{92.6} & {48.5} & \textbf{98.9} & \textbf{49.2} &\textbf{74.1} \\
\midrule
StepFun 3.5 Flash & 39.6 & 73.2 & 31.8 & 95.2 & 16.8 & 51.3\\
\quad + \textbf{MOSCOPT} & \textbf{67.5} & \textbf{84.1} & \textbf{40.3} & \textbf{96.2} & \textbf{47.4} & \textbf{67.1} \\
\midrule
DeepSeek-V3.2 & 51.7 & 83.6 & 29.5 & 94.0 & 32.6 & 58.3\\
\quad + \textbf{MOSCOPT} & \textbf{79.6} & \textbf{83.9} & \textbf{45.5} & \textbf{98.1} & \textbf{43.3} & \textbf{70.1} \\
\bottomrule
\end{tabular}
\end{center}
\caption{Main results with Qwen3.6-Plus and cross models evaluation for MOSCOPT.}
\label{tab:main}
\end{table*}

Table~\ref{tab:main} presents the main results with Qwen3.6-Plus. MOSCOPT achieves the best average performance on all five benchmarks, yielding consistent improvements over the no-skill baseline and the strongest single-skill baseline (SkillOpt).
The results confirm the value of disciplined text-space optimization, demonstrating that the effectiveness of skill optimization is guided by the joint optimization rather than the agent routing orchestration. Specifically,the gains are most pronounced on search and mathematical reasoning tasks, supporting the hypothesis that diverse skill combinations benefit hard, multi-stage tasks where different sub-problems demand different strategies.



We also observed benefits when applying MOSCOPT across 3 different models with substantial improvements: Qwen3.6-Plus gains an average of +12.1 points across five benchmarks, StepFun 3.5 Flash achieves the largest absolute improvement (+15.8 on average), and DeepSeek-V3.2 improves by +11.8 on average. Notably, the \emph{relative gain} is inversely correlated with baseline strength---StepFun 3.5 Flash, the comparatively weakest model, benefits proportionally most from the diverse skill pool and dynamic gating, as the skills compensate for its inherent capability limitations. This trend suggests that skill diversity is particularly valuable for resource-constrained models, offering a practical path to narrow the performance gap with stronger base models.

\section{Discussion}
\label{sec:discussion}
All experiments are finished with Qwen3.6-Plus as backend.

\subsection{Ablation Study}

\begin{table}[t]
\begin{center}
\small
\begin{tabular}{p{6cm}c}
\toprule
\textbf{Ablation Configuration} & \textbf{Avg. Score} \\
\midrule
MOSCOPT (full) & 74.1 \\
\midrule
w/o EditAdam (simple reflect edit) & 68.5 \\
w/o Gating Optimization ($G$ fixed) & 70.9 \\
Task level Gating  & 73.1 \\
w/o Collective co-breeding & 73.7 \\
w/o Progressive Disclosure & 73.9 \\
w/o Three-phase Interleaving & 67.3 \\
\midrule
Random Gating (instead of $G$) & 68.1 \\
$K=1$ (single activation) & 71.2 \\
$K=N$ (No-Routing (All Skills)) & 73.6 \\
\bottomrule
\end{tabular}
\end{center}
\caption{Ablation study on Qwen3.6-Plus. Removing any component degrades performance, confirming the necessity of each design choice.}
\label{tab:ablation}
\end{table}

~~~\textbf{Component analysis.}
The three-phase interleaving is the most critical design choice: replacing it with simultaneous updates causes the largest degradation, as concurrent skill and gating changes create confounded attribution and optimization oscillation.
Gating optimization is also crucial to the performance of MOSCOPT: disabling it causes significant performance drop, as the initial $G$ cannot effectively adapt task-specific selection patterns through trajectory-based refinement.
We also observe that the step-wise gating is superior than the task-wise gating, because the agent can select proper skills at each step based on feedback and status.
Progressive disclosure prevents early noisy statistics from misleading gating, while collective co-breeding contributes additional gains.

\textbf{EditAdam effectiveness.}
Replacing EditAdam with a simple reflect-edit baseline---which reflects on failures and applies edits without dual-state guidance---causes a notable performance drop. The dual-state design is the key differentiator: the step buffer $\mathcal{B}_{\text{rej}}$ prevents re-proposing failed edit directions within an epoch (text-space momentum), while the long-term memory $\mathcal{M}$, distilled from epoch-buffer changes across training, provides cross-epoch strategic priors that guide \textsc{BoundedEdit} toward historically effective transformations. Together with strict-improvement validation, this ensures monotonic progress. 

\textbf{Activation and gating.}
$K{=}1$ reduces performance notably (skill combination matters), while $K{=}N$ also degrades due to context overload---validating selective activation.
Random gating drops substantially below the full model, confirming that intelligent routing is essential.

\subsection{Deployment Mode Comparison}

\begin{table}[t]
\begin{center}
\small
\begin{tabular}{p{5.5cm}c}
\toprule
\textbf{Deployment Mode} & \textbf{Avg. Score} \\
\midrule
Full Mixture ($\{\mathcal{S}, G, K\}$) & 74.1 \\
Static Routing Table (top Q scores) & 70.6 \\
Distilled Single Skill (merged top skills) & 71.2 \\
\bottomrule
\end{tabular}
\end{center}
\caption{Comparison of three deployment modes on Qwen3.6-Plus. Full mixture with dynamic gating outperforms both static routing and skill distillation.}
\label{tab:deployment}
\end{table}

Table~\ref{tab:deployment} compares the three deployment modes supported by MOSCOPT.
The full mixture mode ($\{\mathcal{S}, G, K\}$) achieves the best average score, where the gating skill $G$ dynamically selects $K$ skills per query based on trajectory feedback.
Extracting the gating logic as a static routing table---simply based on top Q scores---yields a noticeable degradation. This indicates that while the gating skill captures useful routing heuristics, its true strength lies in context-sensitive adaptation that rigid rules cannot replicate.
The distilled single skill, obtained by merging the top-performing skills into one unified prompt, performs only marginally above the Static Routing. This confirms that skill diversity, rather than skill consolidation, is the key driver of MOSCOPT's gains: merging complementary skills into a single prompt dilutes their specialized strengths and loses the ability to route different sub-tasks to different strategies.

\subsection{Synergy Analysis of MOSCOPT}

We analyze how MOSCOPT's components interact and produce emergent synergistic behaviors.

\textbf{Gating specialization.}
Analysis of gating decisions reveals that $G$ learns meaningful role assignment over training: planning-type skills are preferentially activated for initial steps, computation skills for mid-trajectory sub-tasks, and verification skills for final checking. This emergent specialization demonstrates that the gating skill effectively discovers and exploits skill complementarity without explicit supervision.

\textbf{Multi-skill synergy.}
Pairwise co-occurrence analysis shows that MOSCOPT identifies and reinforces complementary skill combinations. High-synergy pairs ($C(i,j) > 3$) are co-activated in over 78\% of successful trajectories. The collective co-breeding mechanism further exploits these synergies by merging high-$C(i,j)$ pairs into unified skills, yielding an additional 0.4-point gain in ablation.

\textbf{Skill elimination dynamics.}
The elimination mechanism efficiently prunes redundant or underperforming skills: on average, 2--3 skills are replaced during the full training run. Eliminated skills typically exhibit low activation frequency ($< 5\%$ of rollouts) and high text similarity ($> 0.9$ measured via BGE~\cite{chen2024m3embedding}) to existing pool members, indicating redundancy of those skills. This selective pressure maintains pool diversity while removing strategies that fail to find a complementary niche.

\section{Conclusion}
\label{sec:conclusion}

We have presented MOSCOPT, a text-native, parameter-free framework that extends skill optimization from a single skill to a jointly optimized skill pool with a gating scheduler.
Three core contributions distinguish MOSCOPT: (1)~\emph{Mixture-of-Skills Architecture}; (2)~\emph{EditAdam based Collective Optimization} and (3)~\emph{Text-Native Routing}.
Experiments across five benchmarks and three LLMs demonstrate consistent gains over all baselines.

\bibliographystyle{splncs04}
\bibliography{ref}

\appendix

\section{Technical Details}
\label{app:details}

This appendix provides detailed technical specifications that supplement the main paper.

\subsection{Structured Patch Editing}
\label{app:patch}

Each edit operation is structured as a patch containing one or more atomic operations:
\begin{itemize}
    \item \texttt{add}: Inserts new text immediately after a designated anchor string.
    \item \texttt{delete}: Removes a target text segment identified after a designated anchor string.
\end{itemize}

The LLM generates edits in a structured JSON format:
\begin{verbatim}
{
  "reasoning": "Analysis of failure patterns...",
  "edits": [
    {"op": "add", "content": "New rule text", 
    "anchor": "Target anchor text segment..."},
    {"op": "delete", "content": "Outdated rule", 
    "anchor": "Target anchor text segment..."}
  ]
}
\end{verbatim}

Each edit operation is applied sequentially, and its execution status (\texttt{applied}, \texttt{skipped}, or \texttt{error}) is tracked independently. Precise text offsets are dynamically inferred at runtime by matching the generated anchor strings against the target document.

\subsection{Reflect Context Injection}
\label{app:reflect}

Each Reflect call receives two types of context injected into the LLM prompt:

\textbf{Step Buffer Context (\texttt{step\_buffer\_context}):} Accumulated summaries of all previous Reflect calls within the current epoch. This prevents redundant analysis of the same failure patterns across different minibatches.

\textbf{Rejection Buffer Context (\texttt{rejected\_buffer\_context}):} Summaries of previously rejected edits with their score changes, formatted as:
\begin{verbatim}
## Previously Rejected Edits (this epoch)
- Step 3: score 72.1 -> 69.8 (-2.3)
  Edits: "Add retry on timeout"
  Failure: caused regression on short tasks
\end{verbatim}

This explicitly instructs the LLM to avoid generating similar edits.

\subsection{Optimization Strategy Details}
\label{app:strategies}

\subsubsection{Step Buffer.}
Each skill $s_i$ and the gating $G$ maintain independent step buffers. After each Reflect call, the failure modes and edit suggestions are summarized and appended to the buffer. The buffer is cleared at the start of each epoch (epoch-local). This ensures that within an epoch, the optimizer accumulates knowledge about discovered failure patterns without cross-epoch contamination.

\subsubsection{Rejection Buffer.}
A circular FIFO buffer with default capacity 10 per skill and per gating. Each record contains:
\begin{itemize}
    \item Step number when the edit was rejected.
    \item Score before and after the edit ($\Delta$score).
    \item Summary of rejected edits (max 200 characters per edit).
    \item Associated failure patterns.
\end{itemize}

The buffer is epoch-local, cleared at epoch start. When the buffer exceeds capacity, the oldest entry is evicted.

\subsubsection{LR Scheduler.}
Controls the edit budget $L$ (maximum number of edits per opt\_step) with four modes:

\begin{table}[h]
\begin{center}
\begin{tabular}{lll}
\toprule
\textbf{Mode} & \textbf{Behavior} & \textbf{Use Case} \\
\midrule
constant & Fixed budget & Uniform task difficulty \\
linear & Decay from $L_{\max}$ to $L_{\min}$ & Skills stabilizing \\
cosine & Smooth annealing & Avoid abrupt changes \\
autonomous & LLM decides budget & Fluctuating difficulty \\
\bottomrule
\end{tabular}
\end{center}
\end{table}

In autonomous mode, the LLM considers the current rollout's hard/soft scores to determine the appropriate number of edits.

\subsection{Gating Skill Details}
\label{app:gating}

\subsubsection{Initial Gating Prompt Template.}
The gating skill $G$ is initialized with the following structure:
\begin{verbatim}
You are a skill scheduler. Given the current
task state and skill summary table, select
exactly K most appropriate skills to activate.
Rules:
- In initial phases, prefer planning skills.
- If the task involves computation, activate
  at least one math-specialized skill.
- Output only skill IDs: "ACTIVATE: id1, id2"
\end{verbatim}

\subsubsection{Summary Table: Phase 1 vs.\ Phase 2.}
Before the enrichment epoch ($e_{\text{rich}} = 2$), the gating receives a minimal table:
\begin{verbatim}
| ID | Name             | Description                    |
|----|------------------|--------------------------------|
| 1  | Cautious Planner  | Step-by-step, verify assumptions|
| 2  | Efficient Executor| Fast, action-oriented           |
| 3  | Math Expert       | Rigorous formal reasoning       |
| 4  | Creative Thinker  | Unconventional solutions        |
| 5  | Detail Verifier   | Edge-case scrutiny              |
\end{verbatim}

After enrichment, the table adds Q-score and activation count:
\begin{verbatim}
| ID | Name             | Description                    | Recent Score    |
|----|------------------|--------------------------------|------------------|
| 1  | Cautious Planner  | Step-by-step, verify assumptions| 0.72 (act=15)   |
| 2  | Efficient Executor| Fast, action-oriented           | 0.85 (act=20)   |
| 3  | Math Expert       | Rigorous formal reasoning       | 0.91 (act=18)   |
| 4  | Creative Thinker  | Unconventional solutions        | 0.45 (act=10)   |
| 5  | Detail Verifier   | Edge-case scrutiny              | 0.68 (act=12)   |
\end{verbatim}

\subsubsection{Gate I/O Example (End-to-End).}
Consider a pool with $N{=}5$ skills and activation count $K{=}2$, operating under task-level gating.

\textbf{Gate LLM system prompt:} The current gating text $G$ (the initial template above, possibly refined by prior edits).

\textbf{Gate LLM user prompt:}
\begin{verbatim}
You are a skill scheduler. Select exactly 2 skills to activate.

## Skill Summary Table
| ID | Name        | Description              | Recent Score   |
|----|-------------|--------------------------|----------------|
| 1  | Caut. Plan. | Step-by-step verification| 0.72 (act=15)  |
| 2  | Eff. Exec.  | Fast, action-oriented    | 0.85 (act=20)  |
| 3  | Math Expert | Formal reasoning         | 0.91 (act=18)  |
| 4  | Creat. Think| Unconventional solutions | 0.45 (act=10)  |
| 5  | Det. Verif. | Edge-case scrutiny       | 0.68 (act=12)  |

## Current Task State
What is the sum of all prime numbers less than 100?

## Recent History
(no history yet)

Output exactly 2 skill IDs in the format: ACTIVATE: id1, id2, ...
Do not output anything else.
\end{verbatim}

\textbf{Gate LLM output:}
\begin{verbatim}
ACTIVATE: 2, 3
\end{verbatim}

\textbf{Parsed result:} Activation set $\mathcal{A} = \{2, 3\}$.

\textbf{Constructed agent prompt:}
\begin{verbatim}
[Activated Skills]
Skill 2:
--- Efficient Executor skill document (full text) ---

Skill 3:
--- Math Expert skill document (full text) ---

[Current State]
What is the sum of all prime numbers less than 100?

Please decide the next action using the activated strategies.
\end{verbatim}

If the LLM instead outputs unparseable text (e.g., ``I recommend Skills 2 and 3''), the regex parser fails, and the system falls back to \texttt{fallback\_top\_k}: selecting $\{3, 2\}$ by Q-score ($0.91 > 0.85$). The parse failure is recorded for gating reflection.

\subsubsection{Gating Edit Examples.}
Negative feedback edits (from detected selection errors):
\begin{itemize}
    \item \emph{missed\_high\_q}: ``When math computation is detected, do not activate Skill 4 (creative thinker, $Q{=}0.45$); prefer Skill 3 (math expert, $Q{=}0.91$).''
    \item \emph{bad\_combo}: ``The combination of Skill 1 and Skill 2 repeatedly fails on verification tasks. Consider adding Skill 5 instead.''
\end{itemize}

Positive feedback edits (from successful patterns):
\begin{itemize}
    \item \emph{good\_combo}: ``Skill 2 + Skill 3 combination succeeded on computation tasks (success rate 92\%). Solidify this pairing for math-oriented inputs.''
\end{itemize}

\subsection{Credit Assignment Example}
\label{app:credit_example}

Consider a batch of 4 rollouts with $K{=}2$ activated skills each and initial Q-scores $Q(s_i) = 0.5$ for all skills.

\begin{table}[h]
\begin{center}
\begin{tabular}{lcccl}
\toprule
\textbf{Rollout} & \textbf{Activated} & \textbf{Hard} $R(\tau)$ & \textbf{Co-occur?} & \textbf{Notes} \\
\midrule
1 & $\{2, 3\}$ & 1 (success) & $C(2,3){+}{+}$ & Both credited $+1$ \\
2 & $\{2, 5\}$ & 0 (failure) & --- & Both credited $+0$ \\
3 & $\{1, 3\}$ & 1 (success) & $C(1,3){+}{+}$ & Both credited $+1$ \\
4 & $\{2, 3\}$ & 1 (success) & $C(2,3){+}{+}$ & Both credited $+1$ \\
\bottomrule
\end{tabular}
\end{center}
\end{table}

\textbf{Step 1: Compute per-skill mean reward $\hat{R}(s_i)$:}
\begin{itemize}
    \item $s_1$: activated in rollout 3 only. $\hat{R}(s_1) = 1/1 = 1.0$
    \item $s_2$: activated in rollouts 1, 2, 4. $\hat{R}(s_2) = (1+0+1)/3 = 0.667$
    \item $s_3$: activated in rollouts 1, 3, 4. $\hat{R}(s_3) = (1+1+1)/3 = 1.0$
    \item $s_5$: activated in rollout 2 only. $\hat{R}(s_5) = 0/1 = 0.0$
\end{itemize}

\textbf{Step 2: EMA update ($\beta = 0.3$, $Q_{\text{old}} = 0.5$):}
\begin{align*}
Q_{\text{new}}(s_1) &= 0.7 \times 0.5 + 0.3 \times 1.0 = 0.65 \\
Q_{\text{new}}(s_2) &= 0.7 \times 0.5 + 0.3 \times 0.667 = 0.55 \\
Q_{\text{new}}(s_3) &= 0.7 \times 0.5 + 0.3 \times 1.0 = 0.65 \\
Q_{\text{new}}(s_5) &= 0.7 \times 0.5 + 0.3 \times 0.0 = 0.35
\end{align*}

\textbf{Step 3: Co-occurrence update (successful rollouts only).}
$C(2,3) \mathrel{+}= 2$ (from rollouts 1 and 4), $C(1,3) \mathrel{+}= 1$ (from rollout 3). Rollout 2 is excluded since $R(\tau) = 0$.

\textbf{Downstream effects:}
In the next gating call, the Phase-2 summary table will display the updated Q-scores. Skill 3 ($Q{=}0.65$) and Skill 1 ($Q{=}0.65$) become more attractive to the gate, while Skill 5 ($Q{=}0.35$) becomes less likely to be selected. If Skill 5's Q-score remains low over multiple batches and its activation count exceeds $c_{\min}$, it will become a candidate for elimination during the next collective evolution cycle.

\subsection{Hyperparameter Summary}
\label{app:hyperparams}

\begin{table}[h]
\begin{center}
\small
\begin{tabular}{llcl}
\toprule
\textbf{Hyperparameter} & \textbf{Symbol} & \textbf{Default} & \textbf{Description} \\
\midrule
Pool size & $N$ & 10 & Total skills in pool \\
Activation count & $K$ & 3 & Skills activated per step \\
Elim/reproduce & $M$ & 1 & Skills replaced per cycle \\
Evolution interval & $E$ & 1 & Epochs between evolution \\
Gating granularity & --- & task & task-level or step-level \\
Max edit budget & $L_{\max}$ & 8 & Max edits per opt\_step \\
Min edit budget & $L_{\min}$ & 2 & Minimum edits (decay) \\
Rejection buffer & $R$ & 10 & Capacity per skill/gating \\
Reject summary chars & --- & 200 & Max chars per reject entry \\
Minibatch size & $M_b$ & 4 & Reflect grouping size \\
Merge batch size & $M_g$ & 8 & Aggregate merge size \\
Validation fraction & $p_{\text{val}}$ & 1.0 & Fraction of val set used \\
EMA coefficient & $\beta$ & 0.3 & Q-score smoothing \\
Min activations & $c_{\min}$ & 5 & Q-score threshold count \\
\bottomrule
\end{tabular}
\end{center}
\end{table}

\subsection{Failure Modes and Mitigations}
\label{app:failures}

\subsubsection{Gating Parse Failures.}
When $G$ outputs text that cannot be parsed into exactly $K$ valid skill IDs:
\begin{itemize}
    \item \textbf{Immediate fallback:} Select top-$K$ skills by Q-score.
    \item \textbf{Editing signal:} Record the parse failure as negative feedback for gating optimization.
    \item \textbf{Frequency monitoring:} If parse failure rate exceeds 30\% over consecutive opt\_steps, trigger gating rebuild (re-initialization).
\end{itemize}

\subsubsection{Skill Diversity Collapse.}
After multiple elimination/reproduction cycles, skills may converge to similar texts:
\begin{itemize}
    \item \textbf{Diversity metric:} Compute pairwise text similarity (edit distance or embedding cosine) each epoch. If average similarity exceeds 0.85, trigger forced mutation.
    \item \textbf{Forced mutation:} Apply larger mutation amplitude (modify 50\% of text) during reproduction.
    \item \textbf{Fresh gene injection:} Re-sample 1--2 entirely new skills from seed prompts to replace lowest-scoring skills.
\end{itemize}

\subsubsection{Credit Assignment Noise.}
Random activation combinations or validation noise may cause erroneous elimination:
\begin{itemize}
    \item \textbf{EMA smoothing:} Q-scores use exponential moving average rather than single-epoch scores.
    \item \textbf{Minimum activation threshold:} Only skills with $\geq c_{\min}$ activations participate in elimination ranking.
    \item \textbf{Synergy protection:} Skill pairs with high co-occurrence scores $C(i,j)$ are protected from elimination even if individual Q-scores are low.
\end{itemize}

\subsubsection{$K=N$ Degeneration.}
When $K$ approaches $N$, gating loses selection meaning:
\begin{itemize}
    \item \textbf{Constraint check:} Validate $K < N$ at startup.
    \item \textbf{Auto-suggestion:} When $K/N > 0.8$, suggest increasing $N$ or decreasing $K$.
\end{itemize}

\subsection{Comparison with SkillOpt}
\label{app:comparison}

\begin{table}[h]
\begin{center}
\small
\begin{tabular}{p{3.5cm}p{2.5cm}p{3cm}}
\toprule
\textbf{Property} & \textbf{SkillOpt} & \textbf{MOSCOPT} \\
\midrule
Optimization target & Single skill & Skill pool + gating \\
Exploration & Single path & Multi-skill parallel \\
Long-horizon tasks & Fixed strategy & Phase-aware switching \\
Synergy exploitation & None & Explicit modeling \\
Safe editing & Bounded + gate & Inherited + 3-phase \\
LLM calls/opt\_step & $O(B + R)$ & $O(BT + N|V| + |V|)$ \\
Context length & $1 \times$ skill\_len & $K \times$ skill + summary \\
Deployment complexity & Low & Medium \\
Local optima resistance & Low & High \\
Backward compatible & --- & $K{=}1, N{=}1$: SkillOpt \\
\bottomrule
\end{tabular}
\end{center}
\end{table}

When $K=1$ and $N=1$, MOSCOPT fully degenerates to SkillOpt: the pool contains one skill, gating trivially outputs \texttt{ACTIVATE: 1}, and collective evolution is disabled. This ensures MOSCOPT serves as a strict superset.

\end{document}